# Underwater Doppler Navigation with Self-calibration

Xianfei Pan[1] and Yuanxin Wu[2]

[1](*National University of Defense Technology - College of Mechatronics and Automation, China, 410073)*

[2](*Central South University - School of Aeronautics and Astronautics, China, 410083*)

(E-mail: yuanx_wu@hotmail.com)

Precise autonomous navigation remains a substantial challenge to all underwater platforms. Inertial Measurement Units (IMU) and Doppler Velocity Logs (DVL) have complementary characteristics and are promising sensors that could enable fully autonomous underwater navigation in unexplored areas without relying on additional external Global Positioning System (GPS) or acoustic beacons. This paper addresses the combined IMU/DVL navigation system from the viewpoint of observability. We show by analysis that under moderate conditions the combined system is observable. Specifically, the DVL parameters, including the scale factor and misalignment angles, can be calibrated *in-situ* without using external GPS or acoustic beacon sensors. Simulation results using a practical estimator validate the analytic conclusions.

1. INTRODUCTION. Current underwater navigation technology enables new emerging applications that have been previously considered impossible or impractical, including autonomous naval operations, oceanographic studies and under ice surveys. Despite significant advances, precise navigation remains a substantial challenge to all underwater platforms (Kinsey et al., 2006, Hegrenæs and Berglund, 2009). The Global Positioning System (GPS) provides superior three-dimensional navigation capabilities for vehicles above the water surface, but not underwater due to water blockage of GPS radio-frequency signals. This limits GPS usage to surveying of acoustic transponders or aiding of sensor calibration for underwater applications (Kinsey and Whitcomb, 2006). Acoustic navigation is widely used for limited-area scientific and industrial underwater vehicles, which requires prior careful placement of beacons fixed or moored on the sea floor or on the hull of a surface ship.

The development of commercial Doppler sensors and Inertial Measurement Units (IMU, consisting of a triad of gyroscopes and a triad of accelerometers) has enabled significant improvement to underwater navigation (Kinsey et al., 2006). The multi-beam Doppler Velocity Log (DVL) can provide the bottom-track or water-track velocity measurement with a precision of 0.3% or less, while the IMU measures the arbitrary three-dimensional (3D) angular velocity and translational acceleration which can be integrated to yield navigation information. Combing a DVL with IMU makes possible large-scale underwater navigation in unexplored areas. They are complementary in characteristics to each other. The DVL velocity aids the IMU in mitigating the accumulated navigation errors and in calibrating inertial sensor errors. On the other hand, the IMU's short-time stability helps the DVL to adaptively update parameters that may vary significantly due to ambient factors like water temperature and density (Kinsey et al., 2006).

Their combined navigation accuracy depends on the accuracy of IMU-DVL alignment calibration (Whitcomb et al., 1999; Jalving et al., 2004). The alignment calibration problem is not easy as IMU and DVL are usually separate units and placed at different locations on the vehicle. This mounting arrangement excludes the alignment calibration during manufacture and instead requires an *in-situ* calibration during normal naval operations. Several approaches have been reported to attack this alignment calibration using additional external sensors such as GPS or Long Baseline (LBL) acoustic navigation beacons (Joyce, 1989; Tang et al., 2013; Kinsey and Whitcomb, 2006; 2007). These methods require the vehicle to run on the surface so that a GPS signal is available (Joyce, 1989; Tang et al., 2013), or additional navigation beacons to be placed at surveyed sites (Kinsey and Whitcomb, 2006; 2007). Troni et al. (2012) and Troni and Whitcomb (2010) proposed an *in-situ* calibration method using only on board DVL, gyrocompass and depth sensors, but the gyrocompass is unrealistically assumed to provide exact absolute attitude. In fact, precise attitude estimation for gyrocompass is not a solved problem and also needs extensive study (Silson, 2011; Li et al., 2013). In view of the inherent characteristics of both IMU and DVL, their respective parameter estimation and alignment calibration should be accounted for together, so as to achieve autonomous underwater navigation in large unexplored areas. Otherwise, the combined navigation capacity of IMU and DVL would be compromised in one aspect or another.

The content of the paper is organised as follows. Section 2 presents the mathematical formulation of the combined IMU/DVL navigation system, and then discusses the state estimability from the viewpoint of observability. An ideal observer is derived from the observability analysis procedure. Section 3 designs numerical simulations mimicking typical underwater vehicle motions and uses both the ideal observer and a practical estimator to validate the analytic conclusions. Conclusions are drawn in Section 4.

## 2. SYSTEM FORMULATION AND STATE OBSERVABILITY

2.1. *System Formulation.* The DVL uses the principle that by reflecting three or more non-coplanar radio/sound beams off a surface and measuring the Doppler shifts, the velocity of the body with respect to that surface can be obtained (Groves, 2008). Most systems use the Janus configuration with four beams. The surface-referenced velocity can be obtained as

$$\mathbf{v}^d = \mathbf{K}\Delta\mathbf{f} \tag{1}$$

The superscript $d$ denotes the DVL's coordinate frame $D$ naturally defined by the beam spatial configuration. The vector $\Delta\mathbf{f}$ is formed by Doppler shifts along each beam and the matrix $\mathbf{K}$ depends on the transmitted sound wave frequency, the spatial configuration of beams and the sound speed in water. The first two factors of $\mathbf{K}$ are fixed for a DVL unit, but the sound speed may vary with temperature, depth and salinity by a few percent (Groves, 2008). We rewrite the above relationship as

$$\mathbf{v}^d = \mathbf{K}\Delta\mathbf{f} = \frac{1}{k}\mathbf{K}_s\Delta\mathbf{f} = \frac{1}{k}\mathbf{y}_{DVL} \Leftrightarrow \mathbf{y}_{DVL} = k\,\mathbf{v}^d \tag{2}$$

where $\mathbf{y}_{DVL} \triangleq \mathbf{K}_s\Delta\mathbf{f}$ denotes the DVL output, $k$ is a scalar factor that accounts for the change of sound velocity, and $\mathbf{K}_s$ is a scaled version of $\mathbf{K}$ that is nearly invariant to water conditions.

In contrast to the DVL directly providing the velocity relative to the seabed (bottom-lock), the IMU measures the angular velocity and non-gravitational acceleration with respect to the inertial space. Numerical integrations must be carried out to derive attitude, velocity or position information, which is known as the inertial navigation computation procedure (Groves, 2008; Wu and Pan, 2013b). As gyroscopes and accelerometers are subject to errors like bias and noise, the computed inertial navigation result is prone to error drift accumulating with time.

Denote by $N$ the local level reference frame, by $B$ the IMU body frame, by $I$ the inertial non-rotating frame, and by $E$ the Earth frame. The navigation (attitude, velocity and position) rate equations in the reference n-frame are well known as (Titterton and Weston, 2004; Groves, 2008; Savage, 2007)

$$\dot{\mathbf{C}}_b^n = \mathbf{C}_b^n\left(\boldsymbol{\omega}_{nb}^b \times\right), \quad \boldsymbol{\omega}_{nb}^b = \boldsymbol{\omega}_{ib}^b - \mathbf{b}_g - \mathbf{C}_n^b\left(\boldsymbol{\omega}_{ie}^n + \boldsymbol{\omega}_{en}^n\right), \tag{3}$$

$$\dot{\mathbf{v}}^n = \mathbf{C}_b^n\left(\mathbf{f}^b - \mathbf{b}_a\right) - \left(2\boldsymbol{\omega}_{ie}^n + \boldsymbol{\omega}_{en}^n\right) \times \mathbf{v}^n + \mathbf{g}^n \tag{4}$$

$$\dot{\mathbf{p}} = \mathbf{R}_c \mathbf{v}^n \tag{5}$$

where $\mathbf{C}_b^n$ denotes the attitude matrix from the body frame to the reference frame, $\mathbf{v}^n$ the velocity relative to the Earth, $\boldsymbol{\omega}_{ib}^b$ the error-contaminated body angular rate measured by gyroscopes in the body frame, $\mathbf{f}^b$ the error-contaminated specific force measured by accelerometers in the body frame, $\boldsymbol{\omega}_{ie}^n$ the Earth rotation rate with respect to the inertial frame, $\boldsymbol{\omega}_{en}^n$ the angular rate of the reference frame with respect to the Earth frame, $\boldsymbol{\omega}_{nb}^b$ the body angular rate with respect to the reference frame, and $\mathbf{g}^n$ the gravity vector. The skew symmetric matrix $(\cdot\times)$ is defined so that the cross product satisfies $\mathbf{a}\times\mathbf{b} = (\mathbf{a}\times)\mathbf{b}$ for two arbitrary vectors. The gyroscope bias $\mathbf{b}_g$ and the accelerometer bias $\mathbf{b}_a$ are taken into consideration as approximately random constants, i.e., $\dot{\mathbf{b}}_g = \dot{\mathbf{b}}_a = \mathbf{0}$. The position $\mathbf{p} \triangleq \begin{bmatrix}\lambda & L & h\end{bmatrix}^T$ is described by the angular orientation of the reference frame relative to the Earth frame, commonly expressed as longitude $\lambda$, latitude $L$ and height $h$ above the Earth's surface. In the context of a specific local level frame choice, e.g., North-Up-East, $\mathbf{v}^n \triangleq \begin{bmatrix}v_N & v_U & v_E\end{bmatrix}^T$ and the local curvature matrix is explicitly expressed as

$$\mathbf{R}_c = \begin{bmatrix} 0 & 0 & \dfrac{1}{(R_E + h)\cos L} \\ \dfrac{1}{R_N + h} & 0 & 0 \\ 0 & 1 & 0 \end{bmatrix} \tag{6}$$

where $R_E$ and $R_N$ are respectively the transverse radius of curvature and the meridian radius of curvature of the reference ellipsoid.

Using Equation (2), the derived velocity from IMU is related to the DVL output by

$$\mathbf{y}_{DVL} = k\mathbf{C}_b^d\mathbf{C}_n^b\mathbf{v}^n \tag{7}$$

where $\mathbf{C}_b^d$ is the misalignment attitude matrix of the d-frame with respect to the b-frame. This misalignment matrix and the scale factor are both regarded as random constants. As with previous studies (Hegrenæs and Berglund, 2009; Joyce, 1989; Kinsey and Whitcomb, 2006; 2007; Tang et al., 2013; Troni et al., 2012;, Troni and Whitcomb, 2010), the translational misalignment between DVL and IMU will not be considered hereafter.

We see from Equations (3), (4) and (7) that in addition to attitude/velocity/position that are of immediate interest to us, the combined IMU/DVL navigation also necessitates the finding of such parameters as inertial sensor biases ($\mathbf{b}_g$ and $\mathbf{b}_a$), DVL scale factor $k$ and misalignment matrix $\mathbf{C}_b^d$. Insufficient knowledge of these parameters could result in degrading the combined navigation accuracy.

2.2. *State Observability Analysis*. Without any other external sensors, is it possible to determine the above parameters? From a viewpoint of control system, this question relates to the (global) observability of the system state (Chen, 1999; Wu et al., 2012). In such a case, the system model is given by Equations (3)-(5) and the observation model is given by Equation (7), with the IMU measurement as the system input and the DVL measurement as the system output. Hereafter we use $\mathbf{y}$ to replace $\mathbf{y}_{DVL}$ for notational brevity. From Equation (7), we have $\mathbf{v}^n = \mathbf{C}_b^n \mathbf{C}_d^b \mathbf{y} / k$. Substituting into Equation (4) and using Equation (3) yield

$$\left(\mathbf{C}_b^n\left(\boldsymbol{\omega}_{nb}^b \times\right)\mathbf{C}_d^b\mathbf{y} + \mathbf{C}_b^n\mathbf{C}_d^b\dot{\mathbf{y}}\right)\Big/k = \mathbf{C}_b^n\left(\mathbf{f}^b - \mathbf{b}_a\right) - \left(2\boldsymbol{\omega}_{ie}^n + \boldsymbol{\omega}_{en}^n\right)\times\left(\mathbf{C}_b^n\mathbf{C}_d^b\mathbf{y}\right)\Big/k + \mathbf{g}^n \quad (8)$$

or equivalently,

$$k\left(\mathbf{C}_n^b\mathbf{g}^n + \mathbf{f}^b - \mathbf{b}_a\right) = \left(\left(\boldsymbol{\omega}_{ie}^b + \boldsymbol{\omega}_{ib}^b - \mathbf{b}_g\right)\times\right)\mathbf{C}_d^b\mathbf{y} + \mathbf{C}_d^b\dot{\mathbf{y}} \quad (9)$$

Troni et al. (2012) and Troni and Whitcomb (2010) used a quite coarse simplification of this equation for DVL calibration, for instance, neglecting the term $\mathbf{C}_n^b\mathbf{g}^n$.

On any trajectory segment with constant $\mathbf{C}_n^b$, the quantities $\boldsymbol{\omega}_{ie}^b$ and $\boldsymbol{\omega}_{ib}^b - \mathbf{b}_g$ keep almost unchanged with the moderate speed of an underwater vehicle. As the vector magnitude $\left\|\boldsymbol{\omega}_{ib}^b - \mathbf{b}_g\right\| \approx \left\|\boldsymbol{\omega}_{ie}^b\right\| \approx 7.3\times10^{-5}$ rad/s is very small, Equation (9) is reasonably approximated by

$$k\left(\mathbf{C}_n^b\mathbf{g}^n + \mathbf{f}^b - \mathbf{b}_a\right) = \mathbf{C}_d^b\dot{\mathbf{y}} \quad (10)$$

Time derivative of the above equality is

$$k\dot{\mathbf{f}}^b = \mathbf{C}_d^b\ddot{\mathbf{y}} \quad (11)$$

from which we obtain $k = \pm\left\|\ddot{\mathbf{y}}\right\| / \left\|\dot{\mathbf{f}}^b\right\|$ whenever $\left\|\dot{\mathbf{f}}^b\right\|$ is non-vanishing, as an attitude matrix does not change the magnitude of a vector. It is trivial to solve the sign ambiguity. For example, if the b-frame is roughly aligned with the d-frame, which is often the case in practice, the positive sign will obviously be the right option. So the DVL scale factor is now a known quantity. Note that Equation (11) is valid for any segment of this type, so if there exit two such segments that $\dot{\mathbf{f}}^b$ (or equivalently $\ddot{\mathbf{y}}$) have different directions, the misalignment matrix $\mathbf{C}_d^b$ can be determined according to Lemma 1 in the Appendix.

On any trajectory segment with fast-changing $\mathbf{C}_n^b$, the quantity $\boldsymbol{\omega}_{ib}^b$ is much larger in magnitude than $\boldsymbol{\omega}_{ie}^b$ or $\mathbf{b}_g$ (as far as a quality IMU is concerned). Equation (9) can be approximated as

$$\mathbf{C}_n^b\mathbf{g}^n = \left(\left(\boldsymbol{\omega}_{ib}^b \times\right)\mathbf{C}_d^b\mathbf{y} + \mathbf{C}_d^b\dot{\mathbf{y}}\right)\Big/k - \mathbf{f}^b + \mathbf{b}_a \quad (12)$$

Taking the norm on both sides gives

$$g = \left\|\mathbf{g}^n\right\| = \left\|\boldsymbol{\alpha} + \mathbf{b}_a\right\| \quad (13)$$

which is a quadratic equation on the accelerometer bias $\mathbf{b}_a$ with $\boldsymbol{\alpha} \triangleq \left(\left(\boldsymbol{\omega}_{ib}^b \times\right)\mathbf{C}_d^b \mathbf{y} + \mathbf{C}_d^b \dot{\mathbf{y}}\right)/k - \mathbf{f}^b$. According to Lemma 2 in the Appendix, we know that $\mathbf{b}_a$ will be determined if the vectors $\boldsymbol{\alpha}$, at all times on segments of this type, are non-coplanar, or $\sum \boldsymbol{\alpha}\boldsymbol{\alpha}^T$ is non-singular (Wu et al., 2012). This requirement is naturally met for practical turning in water, as shown in the simulation section.

Then by the chain rule of the attitude matrix, $\mathbf{C}_n^b$ at any time on this segment satisfies

$$\mathbf{C}_n^b = \mathbf{C}_{b(t_0)}^{b(t)} \mathbf{C}_{n(t_0)}^{b(t_0)} \mathbf{C}_{n(t)}^{n(t_0)} \triangleq \mathbf{C}_{b(t_0)}^{b(t)} \mathbf{C}_n^b(t_0) \mathbf{C}_{n(t)}^{n(t_0)} \tag{14}$$

where $\mathbf{C}_n^b(t_0)$ denotes the initial attitude matrix at the beginning of this segment, and $\mathbf{C}_{b(t_0)}^{b(t)}$ and $\mathbf{C}_{n(t_0)}^{n(t)}$, respectively, encode the attitude changes of the body frame and the reference frame. Substituting Equation (14) into Equation (12),

$$\mathbf{C}_n^b(t_0) \mathbf{C}_{n(t)}^{n(t_0)} \mathbf{g}^n = \mathbf{C}_{b(t)}^{b(t_0)} \left(\boldsymbol{\alpha} + \mathbf{b}_a\right) \tag{15}$$

$\mathbf{C}_n^b(t_0)$ is solvable as there always exist two time instants that $\mathbf{C}_{n(t)}^{n(t_0)} \mathbf{g}^n$ have different directions due to the Earth rotation (Wu et al., 2012), so is the attitude matrix $\mathbf{C}_n^b$ using Equation (14). Then the velocity $\mathbf{v}^n$ can be determined by Equation (7), and the gyroscope bias $\mathbf{b}_g$ can be determined by Equation (3).

The above analysis is summarised in the theorem below.

*Theorem 1 (State Observability)*: If the vehicle trajectory contains segments of constant attitude with linearly independent $\dot{\mathbf{f}}^b$ (Type-I), as well as turning segments on which the vectors $\boldsymbol{\alpha}$ as defined in Equation (13) are non-coplanar (Type-II), then the system of Equations (3)-(5) and (7) is observable in attitude, velocity, inertial sensor biases, and DVL scale factor and misalignment matrix.

Here are some explanations on the condition of linearly independent $\dot{\mathbf{f}}^b$, the rate of the specific force (sum of external forces except gravitation).

*Remark 1*: For a multiple-thruster-propelled underwater vehicle as in Kinsey and Whitcomb (2006; 2007), it is not difficult to fulfil this condition, for example by thruster switching. This will normally create driving forces in different directions in the IMU body frame, resulting in linearly independent $\dot{\mathbf{f}}^b$.

*Remark 2*: For an underwater vehicle with a single thruster, it is tricky to fulfil this condition while keeping constant attitude. No matter the thruster force or the water resistance, its direction is fixed relative to IMU or DVL. If the vehicle moves at the same depth, linearly independent $\dot{\mathbf{f}}^b$ might only occur instantaneously at both start and end times of Type-I segments (a situation much like that in Tang et al. (2009)). This excitation is in practice not sufficient to produce a good estimate. Consider an example where $\dot{\mathbf{f}}^b = \begin{bmatrix} \dot{f}_x^b & 0 & 0 \end{bmatrix}^T$ for all Type-I segments, which means the thruster force aligns with *x*-axis of the IMU body frame. Suppose the rotation sequence from d-frame to b-frame is first around *y*-axis (yaw angle, $\psi$), followed by *z*-axis (pitch angle, $\theta$) and then by *x*-axis (roll angle, $\phi$), $\mathbf{C}_d^b$ can be re-parameterised in Euler angles as

$$\mathbf{C}_d^b = \begin{bmatrix} \cos\theta\cos\psi & \sin\theta & -\cos\theta\sin\psi \\ \sin\phi\sin\psi - \cos\phi\cos\psi\sin\theta & \cos\phi\cos\theta & \cos\psi\sin\phi + \cos\phi\sin\theta\sin\psi \\ \cos\phi\sin\psi + \cos\psi\sin\phi\sin\theta & -\cos\theta\sin\phi & \cos\phi\cos\psi - \sin\phi\sin\theta\sin\psi \end{bmatrix}. \quad (16)$$

Substituting into Equation (11)

$$k\,\dot{f}_x^b\left[\cos\theta\cos\psi \quad \sin\theta \quad -\cos\theta\sin\psi\right]^T = \ddot{\mathbf{y}} \quad (17)$$

which is irrelevant to the roll angle $\phi$ around $x$-axis. The other two angles can be computed by element comparison, while the angle $\phi$ is not estimable. This case is right what we encountered in land vehicle navigation subject to the non-holonomic constraint (Wu et al., 2009). It is the angle around the thruster force that is inestimable, so if the thruster force was in a general direction, it can be imagined that all three Euler angles would be affected.

*Remark 3*: Even if $\dot{\mathbf{f}}^b$ (or equivalently $\ddot{\mathbf{y}}$) on the segments of constant attitude are linearly dependent, all states in Theorem 1 are still observable except the angle around the direction of $\dot{\mathbf{f}}^b$. This interesting conclusion can be obtained from the proof of Theorem 1. We only need to show that the products $\mathbf{C}_d^b\mathbf{y}$ and $\mathbf{C}_d^b\dot{\mathbf{y}}$ are known quantities in this scenario, in spite of the undetermined $\mathbf{C}_d^b$. Denote by $\boldsymbol{\eta}$ the unit direction of linearly dependent $\ddot{\mathbf{y}}$. For a single-thruster underwater vehicle, both the DVL velocity $\mathbf{y}$ and the velocity rate $\dot{\mathbf{y}}$ almost certainly align with the direction of $\boldsymbol{\eta}$. From Equation (11), for some time $t_1$ on this segment $k\dot{\mathbf{f}}^b(t_1) = \mathbf{C}_d^b\ddot{\mathbf{y}}(t_1) = \left\|\ddot{\mathbf{y}}(t_1)\right\|\mathbf{C}_d^b\boldsymbol{\eta}$ where $\ddot{\mathbf{y}}(t_1)$ is non-zero. Therefore, $\mathbf{C}_d^b\mathbf{y} = \|\mathbf{y}\|\mathbf{C}_d^b\boldsymbol{\eta} = k\|\mathbf{y}\|\dot{\mathbf{f}}^b(t_1)/\left\|\ddot{\mathbf{y}}(t_1)\right\|$ and $\mathbf{C}_d^b\dot{\mathbf{y}} = \|\dot{\mathbf{y}}\|\mathbf{C}_d^b\boldsymbol{\eta} = k\|\dot{\mathbf{y}}\|\dot{\mathbf{f}}^b(t_1)/\left\|\ddot{\mathbf{y}}(t_1)\right\|$ are both functions of known quantities.

*Remark 4*: Ascending or descending motion makes it possible for a single-thruster underwater vehicle to have linearly independent $\dot{\mathbf{f}}^b$. The water resistance due to ascending/descending velocity change will incur non-zero $\dot{\mathbf{f}}^b$ that deviates in direction from the thruster force. The ascending/descending motions are realised by unloading/loading the ballast tank.

2.3. *Derived Ideal Observers.* The above observability analysis not only tells a ‘yes or no’ answer to state estimability, but provides us valuable insights on designing observers. This is an additional advantage of global observability analysis (Wu et al., 2012). Specifically, Equation (11) allows us to construct an ideal observer to estimate DVL parameters, using IMU and DVL measurements on Type-I segments. No additional external sensors like GPS or LBL beacons are required. Suppose the time interval $\left[t_s^j \quad t_e^j\right]$ corresponds to the j-th Type-I segment ($j = 1, 2, \ldots$). For any $t \in \left[t_s^j \quad t_e^j\right]$, integrating (11) twice over the subinterval $\left[t_s^j \quad t\right]$

$$k\boldsymbol{\beta}(t) = \mathbf{C}_d^b\boldsymbol{\gamma}(t) \quad (18)$$

where $\boldsymbol{\beta}(t) \triangleq \int_{t_s^j}^{t}\mathbf{f}^b d\tau - \mathbf{f}^b\left(t_s^j\right)\left(t - t_s^j\right)$ and $\boldsymbol{\gamma}(t) \triangleq \mathbf{y}(t) - \mathbf{y}\left(t_s^j\right) - \dot{\mathbf{y}}\left(t_s^j\right)\left(t - t_s^j\right)$. Compared with Equation (11), this integral form spares the differentiation of the measurements that would be subject to noise amplification (Wu and Pan, 2013a; Wu et al., 2014). The above equation obviously applies to all Type-I segments. Denote $\Omega = \bigcup_j \left[t_s^j \quad t_e^j\right]$, we have the following observer to calibrate the DVL parameters.

Ideal Observer – DVL Calibration (IO-DVLC):

1) Compute $k = \|\boldsymbol{\gamma}(t)\| / \|\boldsymbol{\beta}(t)\|$ for each $t \in \Omega$;

2) Obtain $\mathbf{C}_d^b$ by solving the constrained optimization problem $\min_{\mathbf{C} \in SO(3)} \int_\Omega \|k\boldsymbol{\beta}(t) - \mathbf{C}\boldsymbol{\gamma}(t)\|^2 dt$ as done in (Wu and Pan, 2013a).

The "ideal observer" will be used to verify the analytic conclusions above.

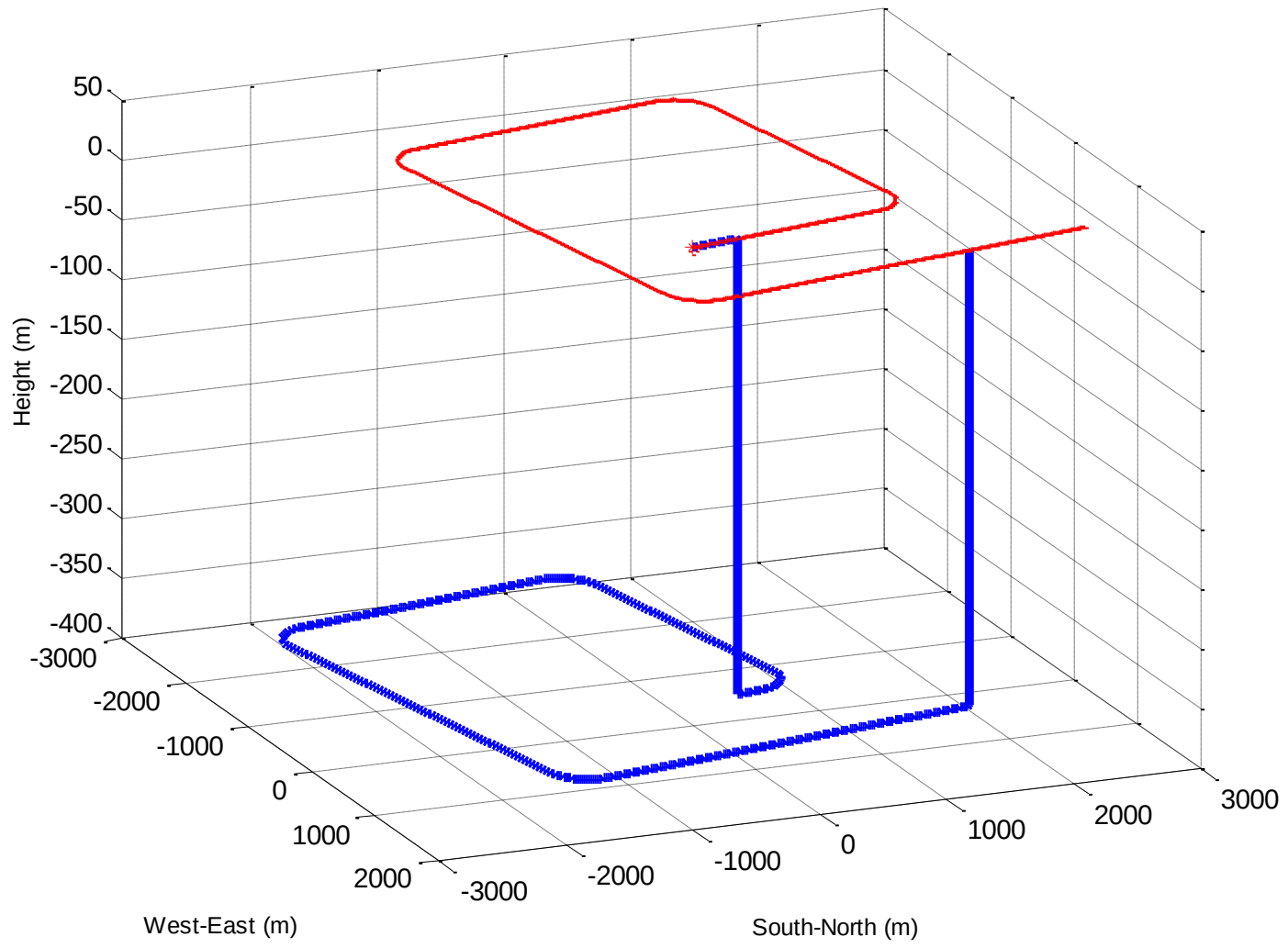

Figure 1. Moving trajectory of the underwater vehicle (3D run: in blue, 2D run: in red).

3. NUMERICAL STUDY. In this section, we carry out numerical simulations to study the validity of the analysis. We will see that the above analysis can explain quite well what we will encounter in practical state observers or estimators. The simulations are designed to mimic the typical motions of a single-thruster propelled underwater vehicle. The 3D trajectory of the vehicle is illustrated in Figure 1 and the motion sequences with time are listed in Table 1. The thruster force is arranged along the vehicle's longitudinal direction (x-axis of the vehicle frame, v-frame, defined as forward-upward-rightward). On board the vehicle is an IMU equipped with a triad of gyroscopes (bias $0.01°/h$, noise $0.1°/h/\sqrt{\text{Hz}}$) and accelerometers (bias $50\mu g$, noise $10\mu g/\sqrt{\text{Hz}}$), and a DVL with measurement noise 2 cm/s ($1\sigma$). The vehicle v-frame is assumed to align perfectly with the IMU (b-frame), from which the DVL (d-frame) misaligns in attitude by -0.1 °(roll), -0.5 °(yaw) and -0.2 °(pitch). The DVL scale factor is set to 0.9998. The initial attitude of the IMU is set to 3 °(roll), 10 °(yaw) and zero (pitch).

Table 1. Motion Sequences with Time.

| Time (s) | | Motions | Segment Type |
|---|---|---|---|
| 3D Run | 2D Run | | |
| 0-600 | 0-600 | Static | / |
| 600-660 | 600-800 | Level motion with constant attitude but varying specific force rate | I |
| 660-720 | / | Descending with constant attitude but varying specific force rate | I |
| 720-750 | / | Level motion with constant attitude but varying specific force rate | I |
| 750-1970 | 800-2040 | Running along a square shape, with tilted turning and constant speed | I & II |
| 1970-2000 | / | Level motion with constant attitude but varying specific force rate | I |
| 2000-2060 | / | Ascending with constant attitude but varying specific force rate | I |

The DVL output profile and IMU output profile generated by 3D motion sequences in Table 1 are respectively plotted in Figures 2 and 3. As the vehicle moves forward over 600-660 s and downward over 660-720 s, and the vectors of specific force rate (Figure 3) or DVL acceleration on these two segments are linearly independent (Figure 2), the condition of Theorem 1 for Type-I segments is satisfied. The derived IO-DVLC observer in Section 2.3 is used to estimate the DVL scale factor (Figure 4) and misalignment angles (Figure 5). We see that the scale factor is computable at either of two segments, while the misalignment angles, especially the roll angle, cannot converge until the downward segment is carried out at 660 s. These observations have been well predicted already by Remarks 2 and 4, although the IO-DVLC observer is not accurate enough in the misalignment angles.

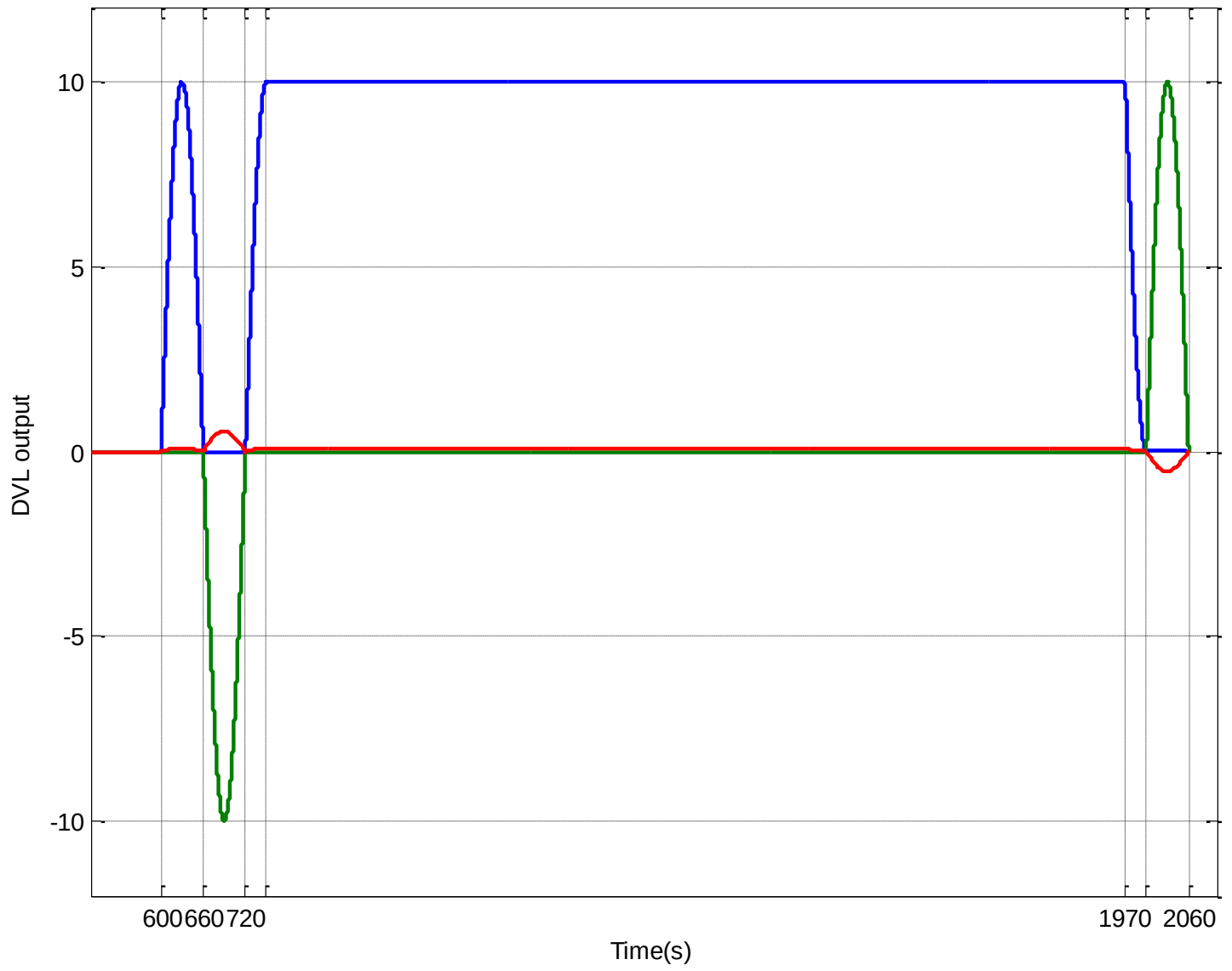

Figure 2. Profile of DVL outputs with specific time tag.

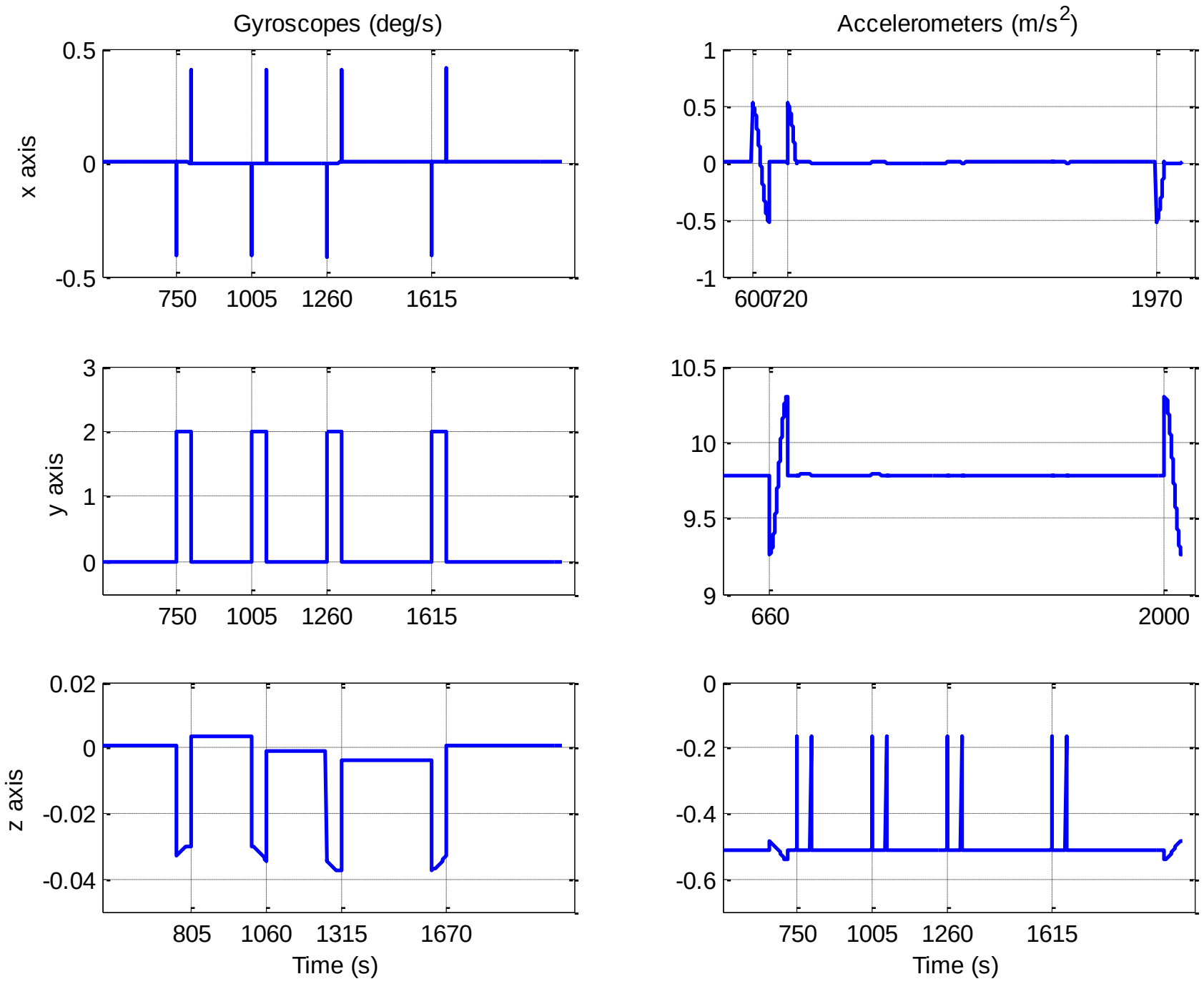

Figure 3. Profile of IMU gyroscope/accelerometer outputs with specific time tag.

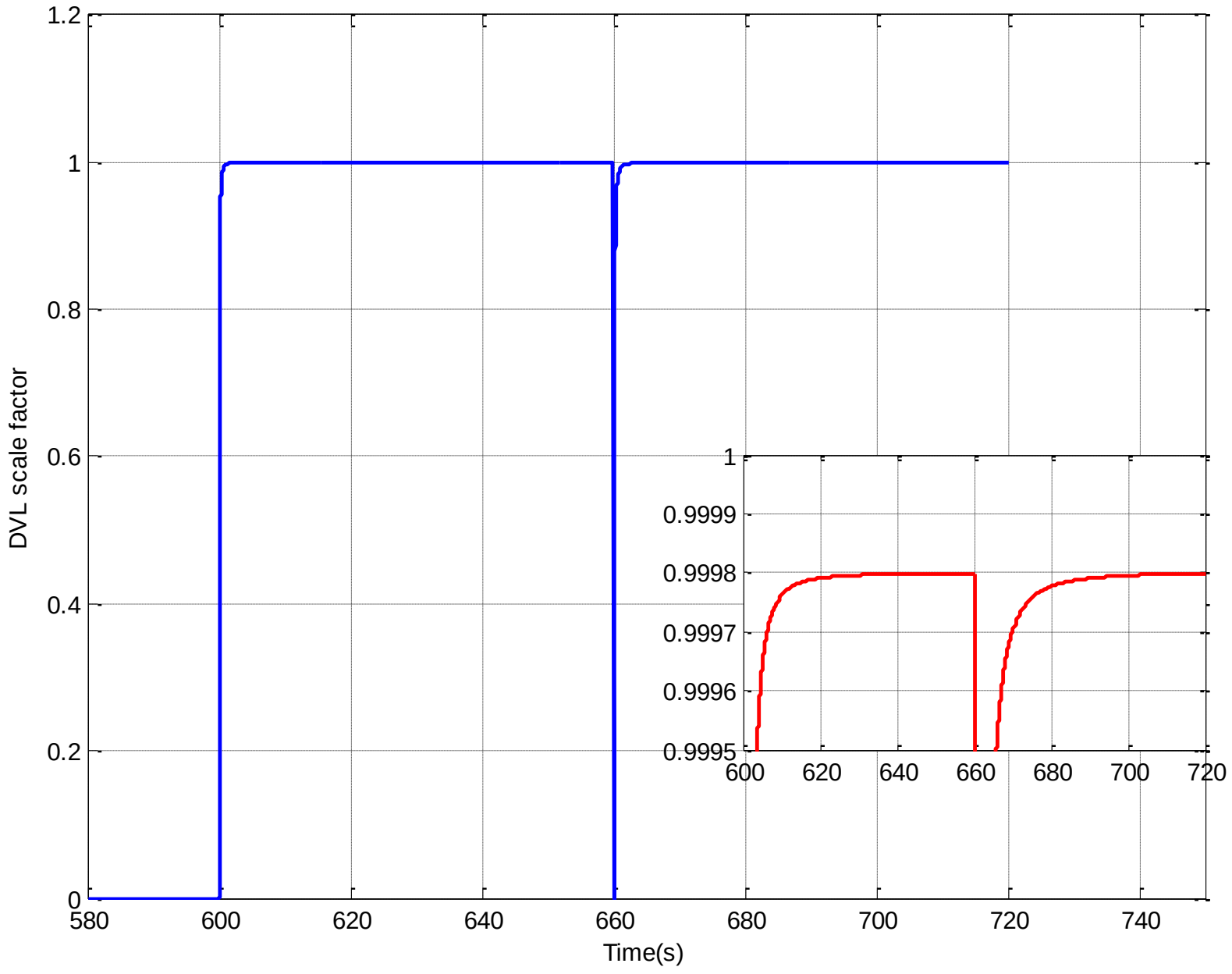

Figure 4. DVL scale factor estimate by ideal observer IO-DVLC.

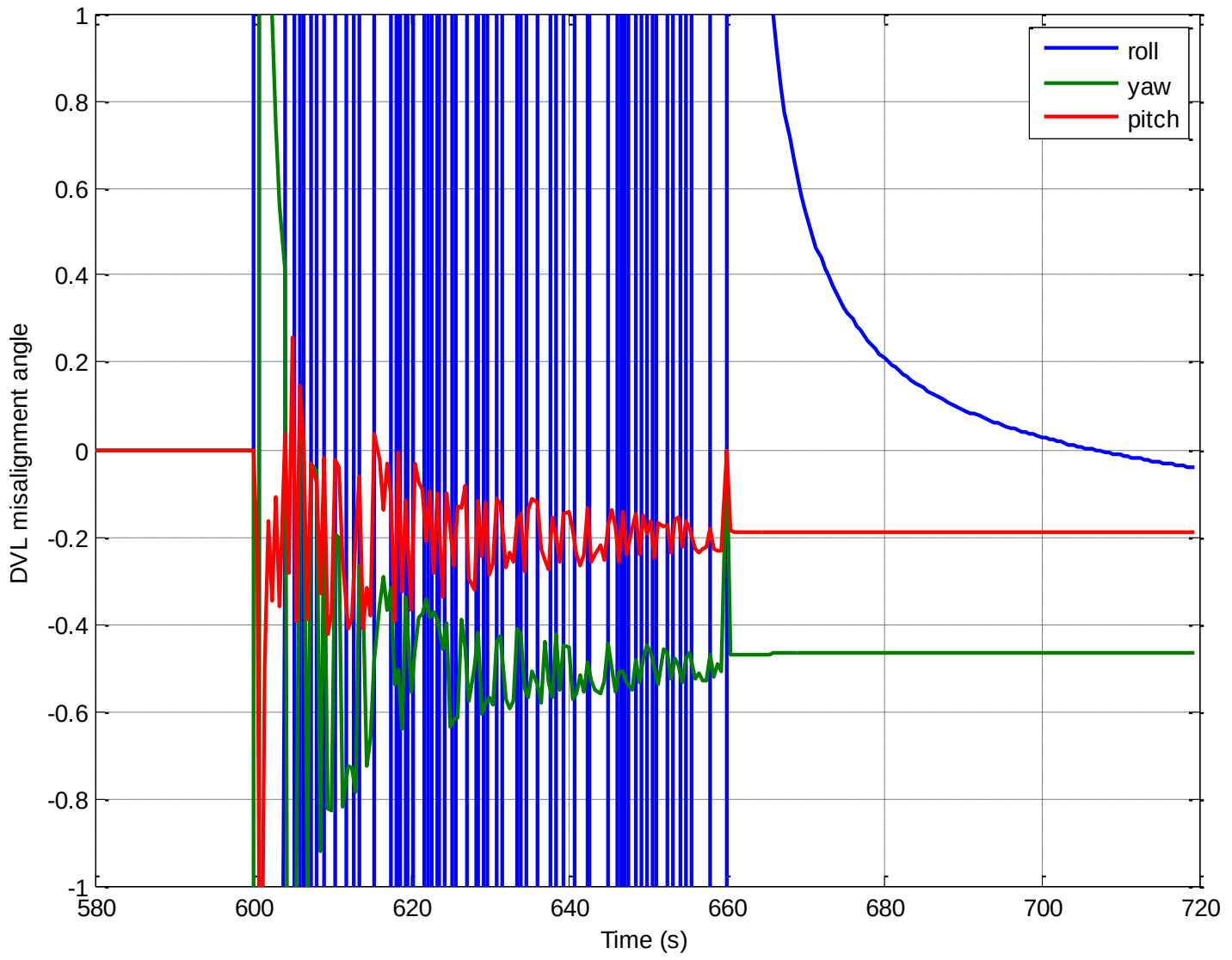

Figure 5. DVL misalignment angle estimate by ideal observer IO-DVLC (unit: degree).

Next we implement an Extended Kalman Filter (EKF) to estimate the states of the combined IMU/DVL system, of which the system dynamics are given by Equations (3)-(5) and the measurement is given by Equation (7). The EKF is a nonlinear state estimator widely used in numerous applications. In addition to those states in Theorem 1, the EKF also estimates the position. The position is unobservable, but it is correlated with other states through the system dynamics and the determination of other states will help mitigate the position error drift. The first 600 s static segment is used for IMU initial alignment. Additional angle errors of 0.1 °(1σ, for yaw) and 0.01 °(1σ, for roll and pitch) are added to the final alignment result so as to mimic the influence of non-benign alignment conditions underwater. Figures 6 and 7 respectively present the DVL scale factor estimate and misalignment angle estimate, as well as their standard variances. The scale factor in Figure 6 converges swiftly from the initial value 0.8 to the truth once the vehicle starts to move at 600 s, as are the two misalignment angles, yaw and pitch, in Figure 7. The roll angle approaches its true value when the descending motion starts at 660 s. Apparently, EKF is more accurate than the IO-DVLC observer in estimating the DVL parameters. The inertial sensor bias estimates and their standard variances are presented in Figures 8 and 9. We see that turning on the square trajectory after 750 s drives the accelerometer bias estimate to convergence (Figure 9). The gyroscope bias in Figure 8 is relatively slower in convergence, especially that in vertical direction (y-axis), due to weaker observability. Figures 10 and 11 give the attitude error and positioning error, as well as their standard variances. The normalised standard variances for attitude, gyroscope/accelerometer biases, DVL scale factor and misalignment angles are plotted in Figure 12. It is the DVL scale factor and misalignment angles (yaw and pitch) that have the strongest observability in this simulation scenario. The interaction among states are quite obvious from Figure 12; for example, yaw angle and the DVL roll angle (at 660 s), and roll/pitch angles and the x-axis accelerometer bias (at 750 s). All of the EKF's behaviours accord with Theorem 1.

A 2D trajectory is also examined to verify the analysis of Remark 3. This planar run is similar with the above 3D run, but excludes the descending/ascending segments (as seen in Figure 1 and Table 1). The DVL scale factor and misalignment angle estimates, attitude error and position error by EKF are respectively plotted in Figures 13-16. As predicted by Remarks 2 and 3, all states but the DVL roll angle are estimable. Although the DVL roll angle does not converge (Figure 14) in this case, other states are estimated quite well.

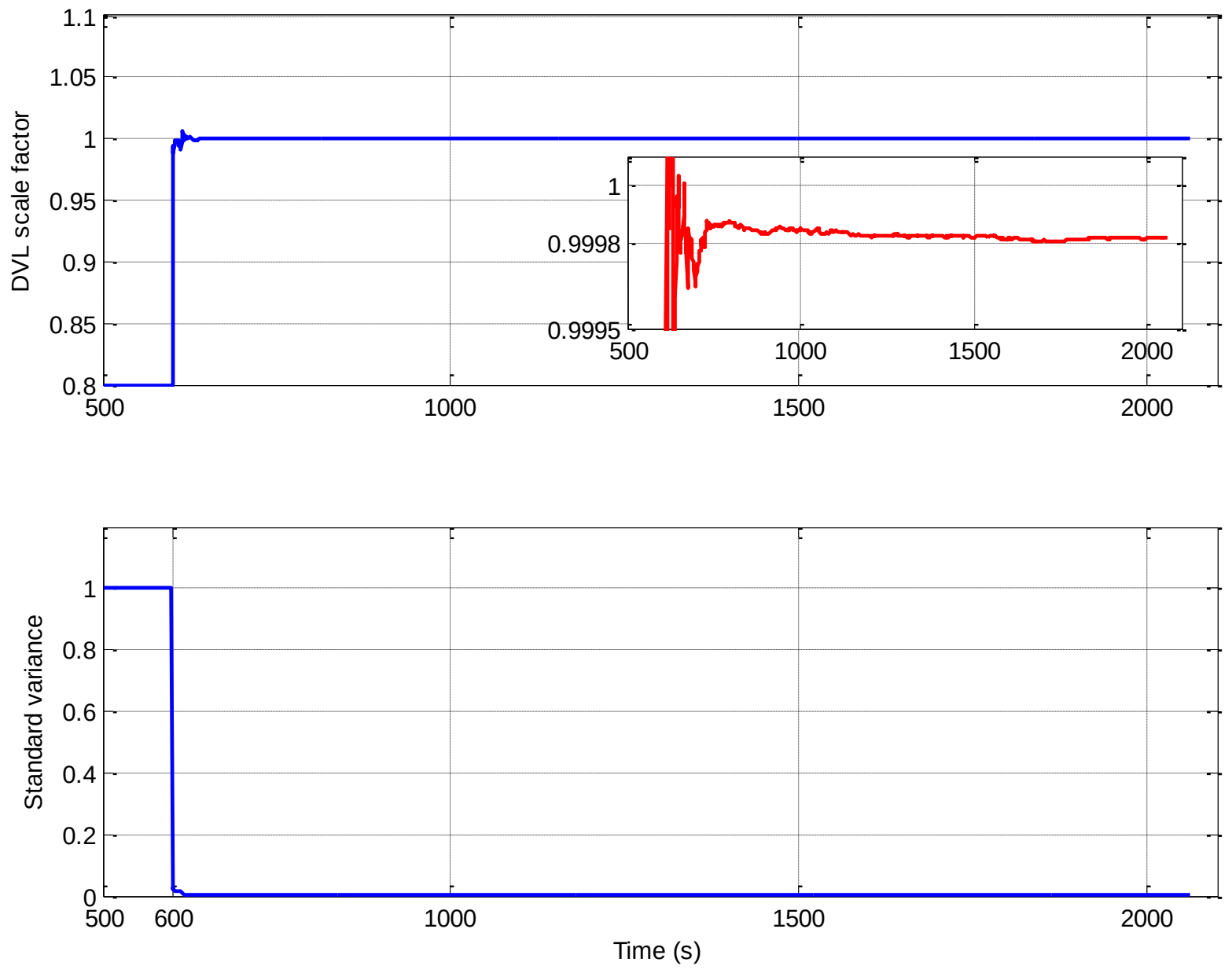

Figure 6. DVL scale factor estimate and standard variance by EKF.

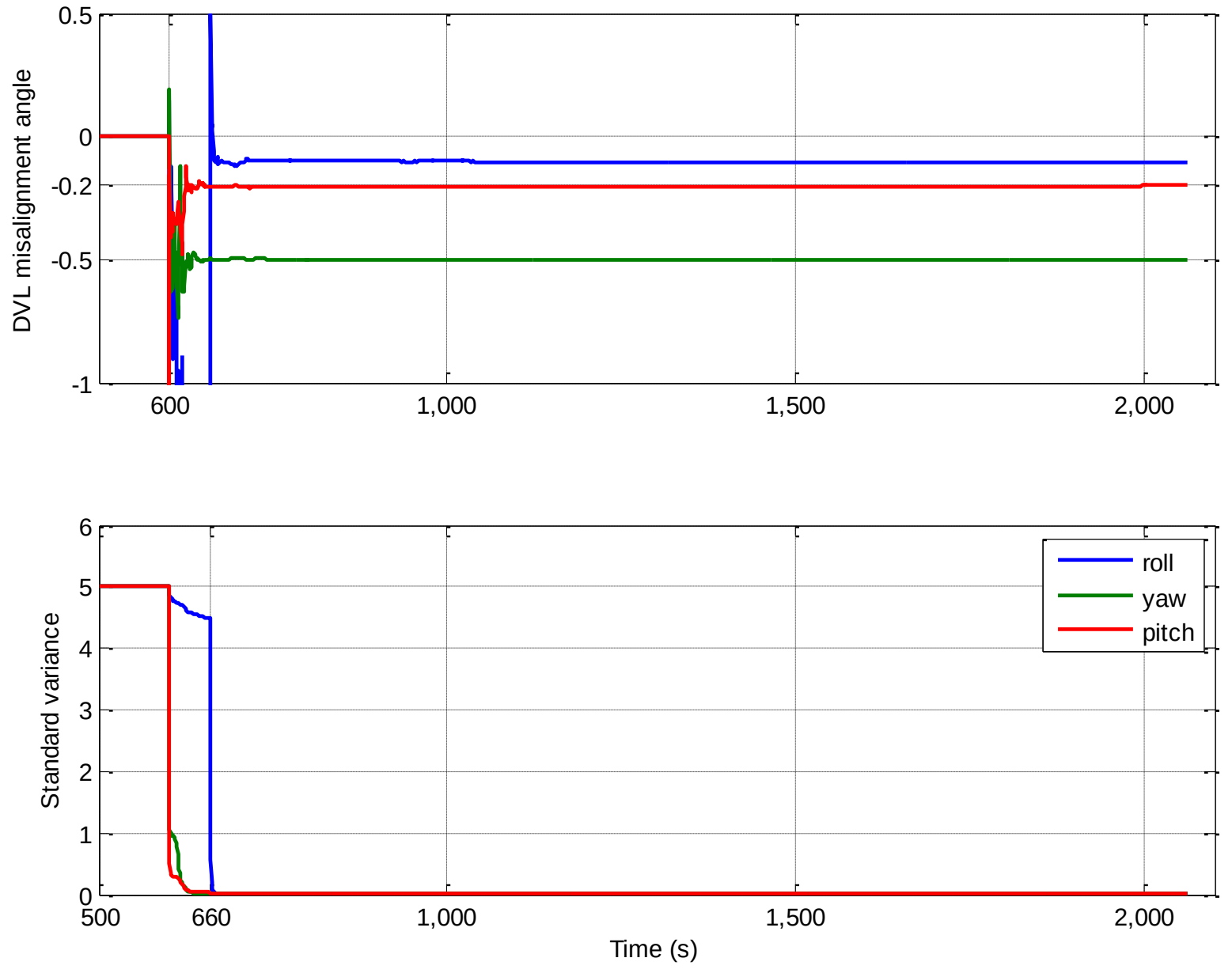

Figure 7. DVL misalignment angle estimate and standard variance by EKF (unit: °).

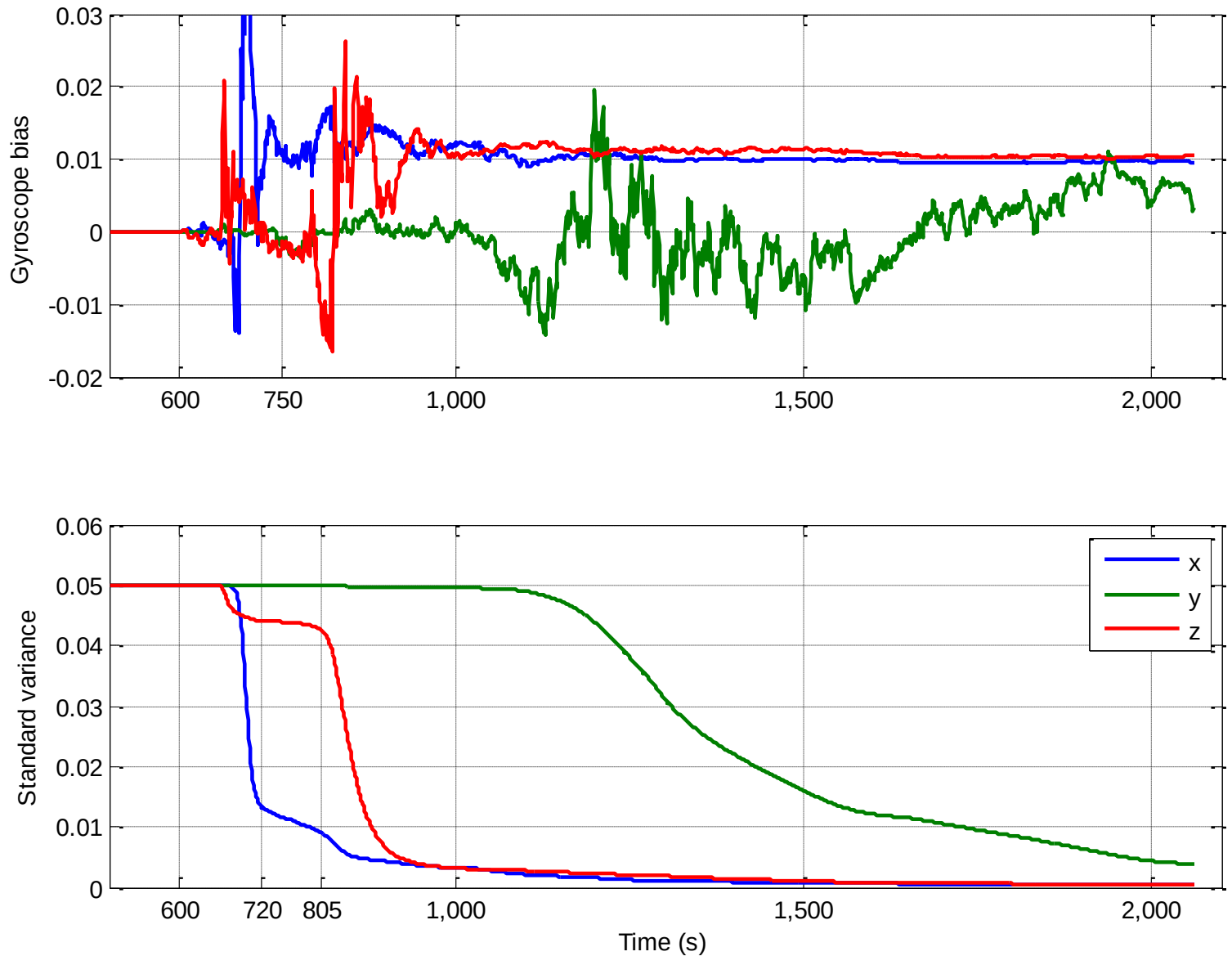

Figure 8. Gyroscope bias estimate and standard variance by EKF (unit: °/h)

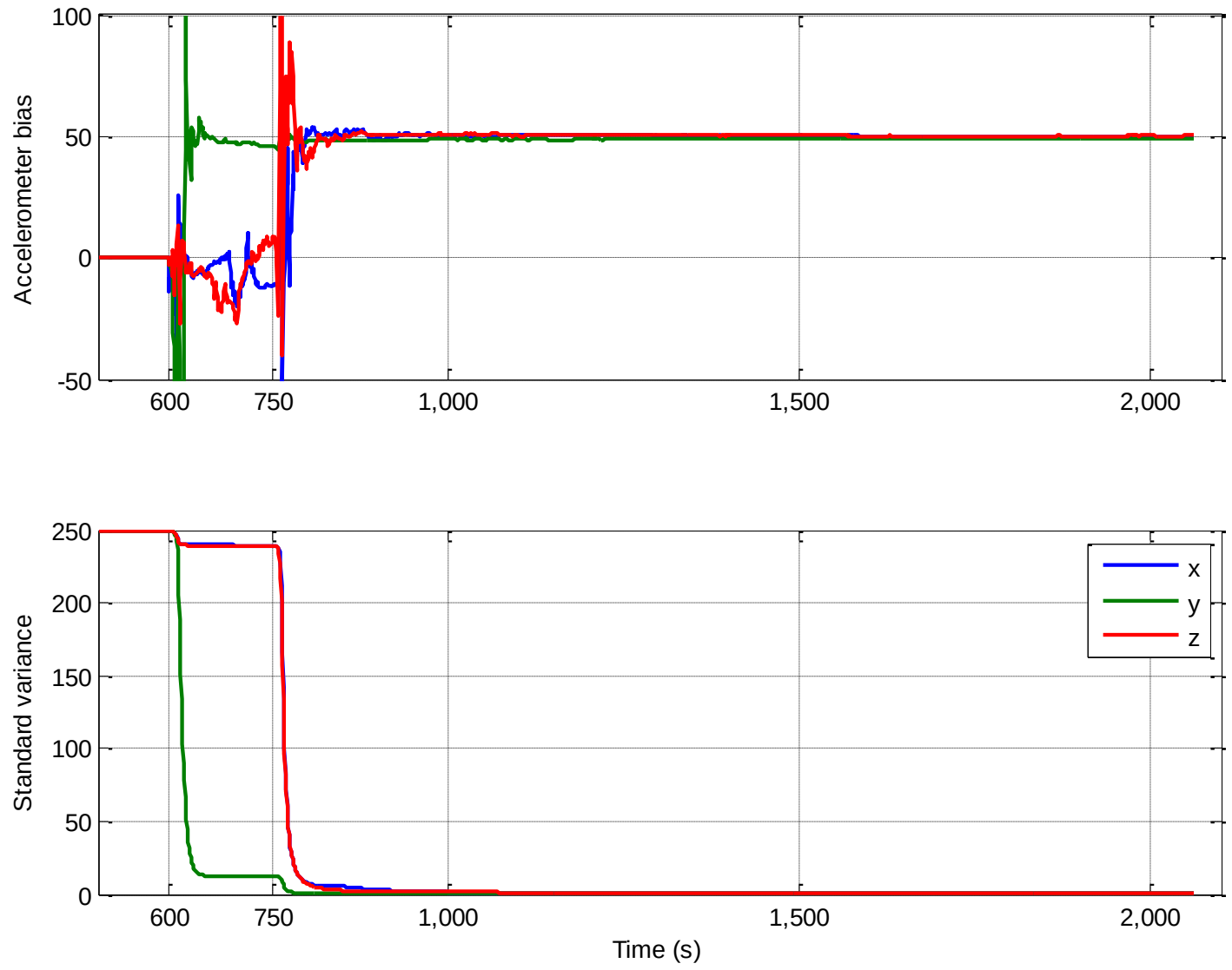

Figure 9. Accelerometer bias estimate and standard variance by EKF (unit: micro g).

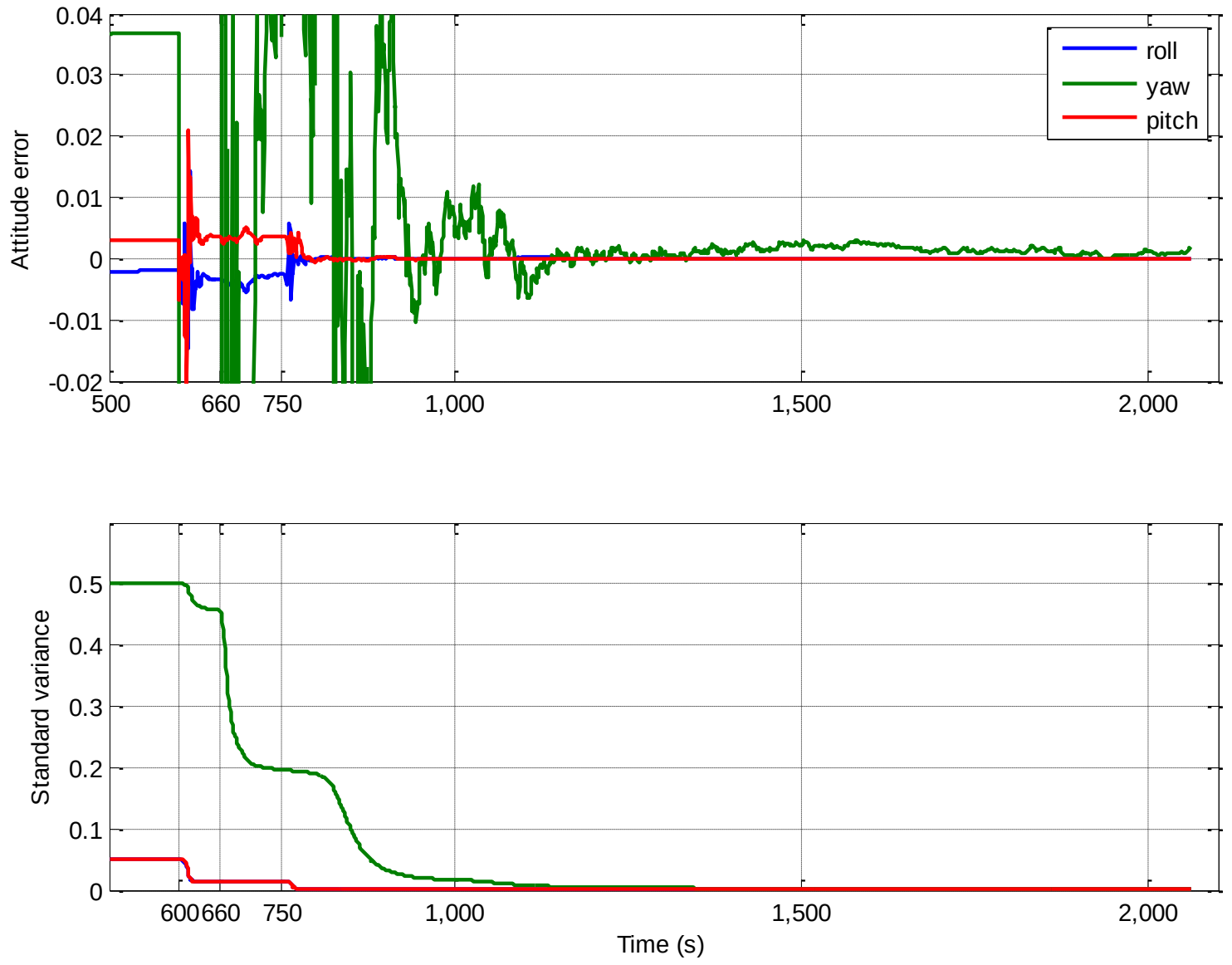

Figure 10. Attitude error and standard variance by EKF (unit: °).

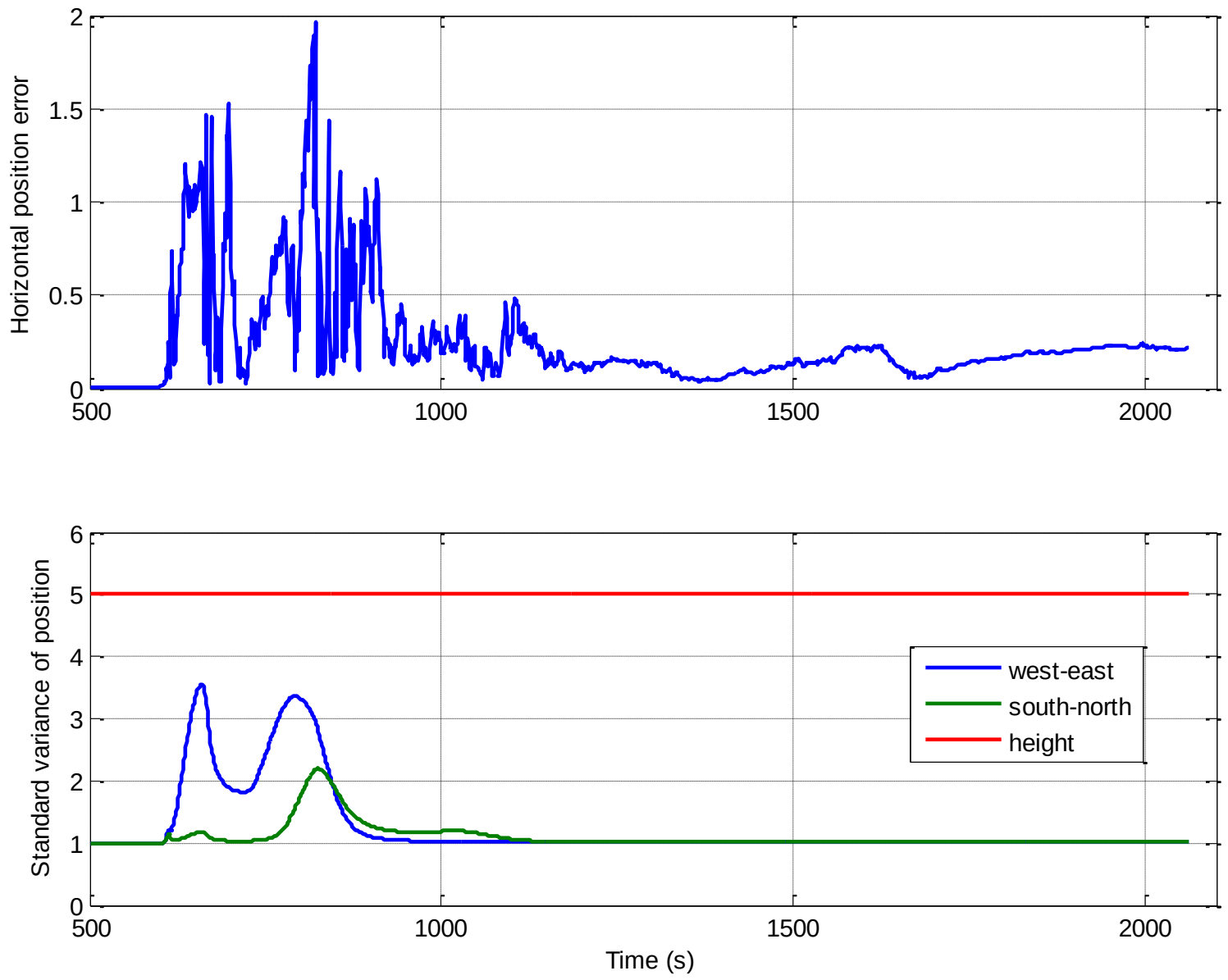

Figure 11. Horizontal position error and standard variance of position estimate by EKF (unit: metre).

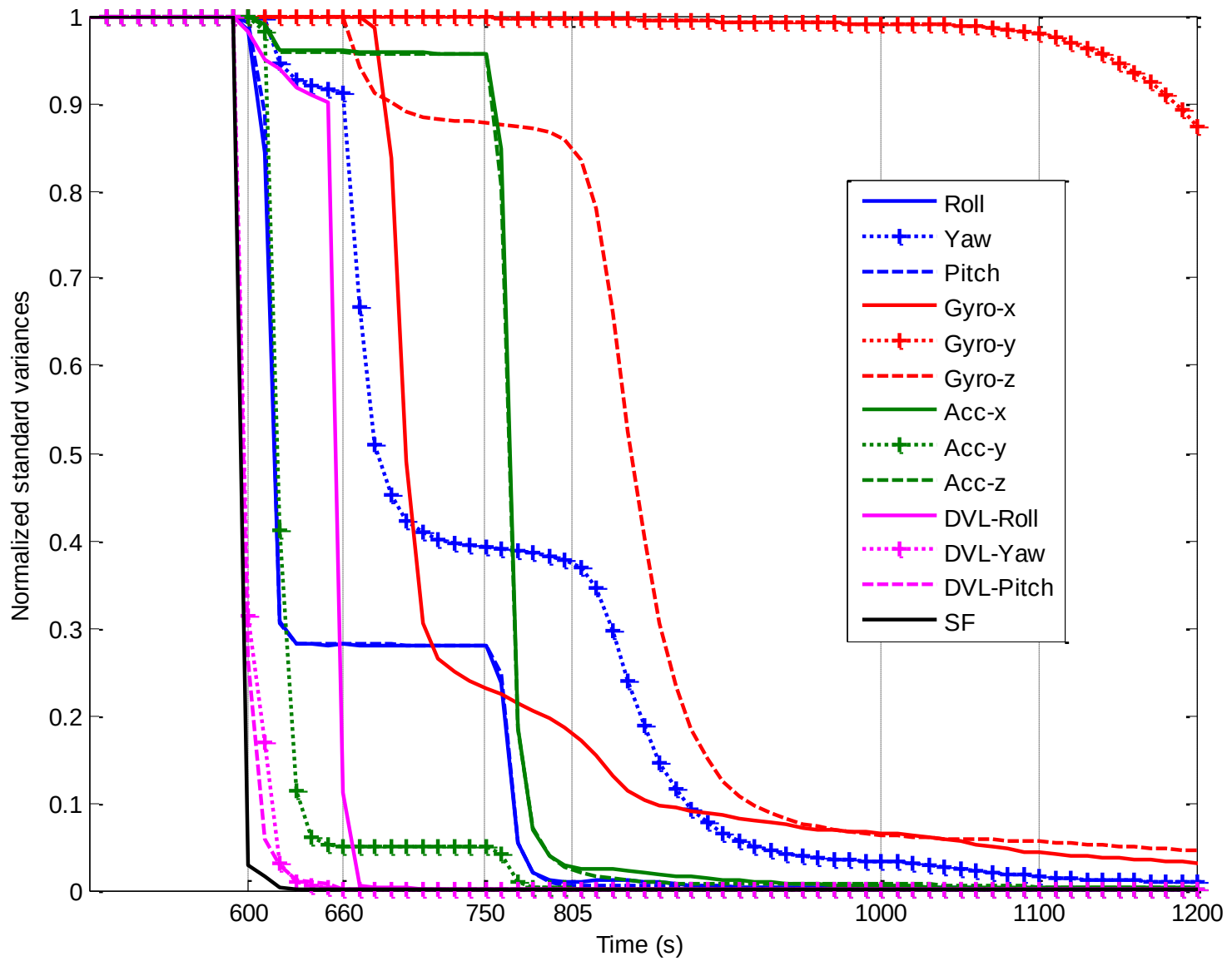

Figure 12. Normalised standard variances for attitude, inertial sensor bias, DVL scale factor and misalignment angles.

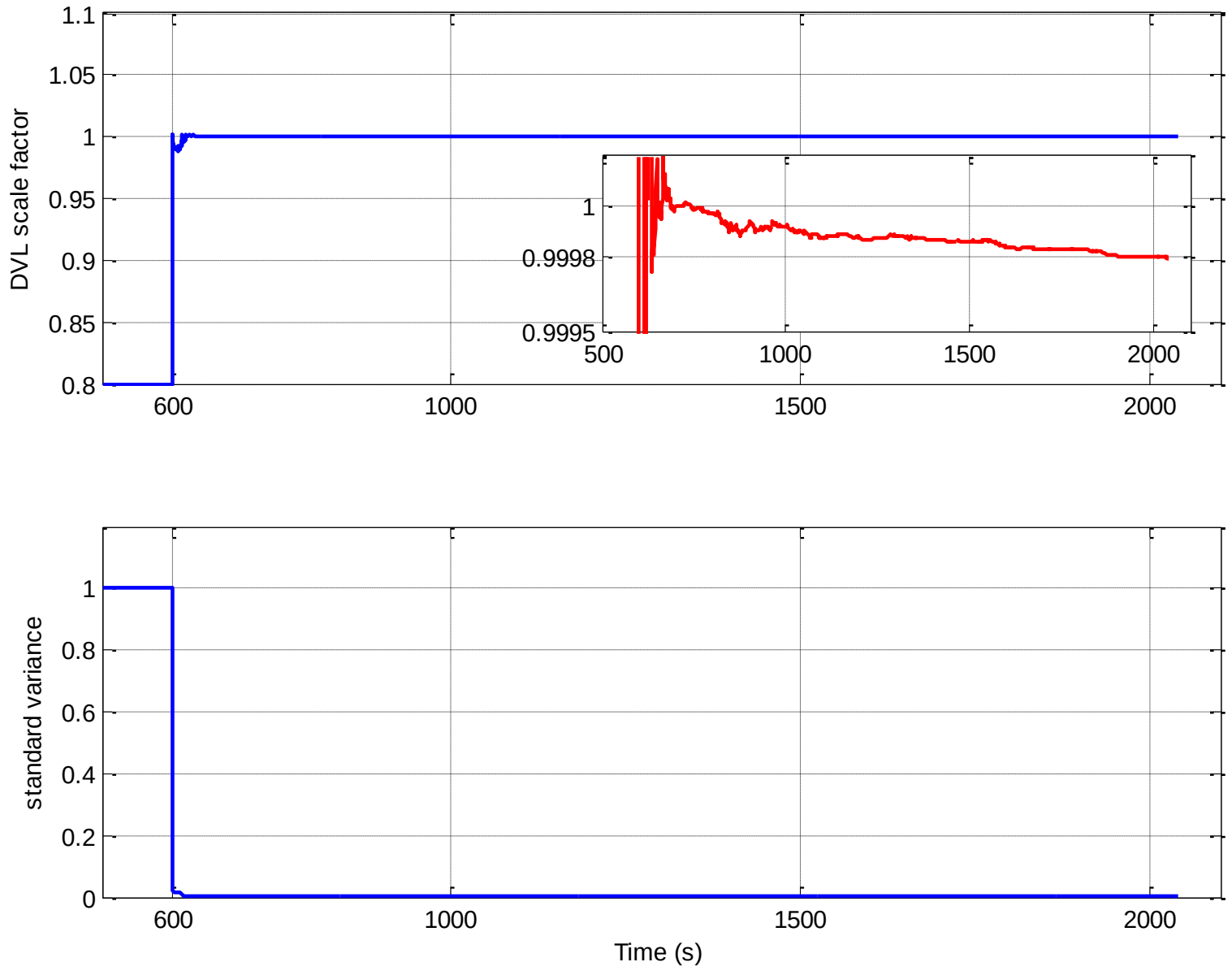

Figure 13. DVL scale factor estimate by EKF in 2D run.

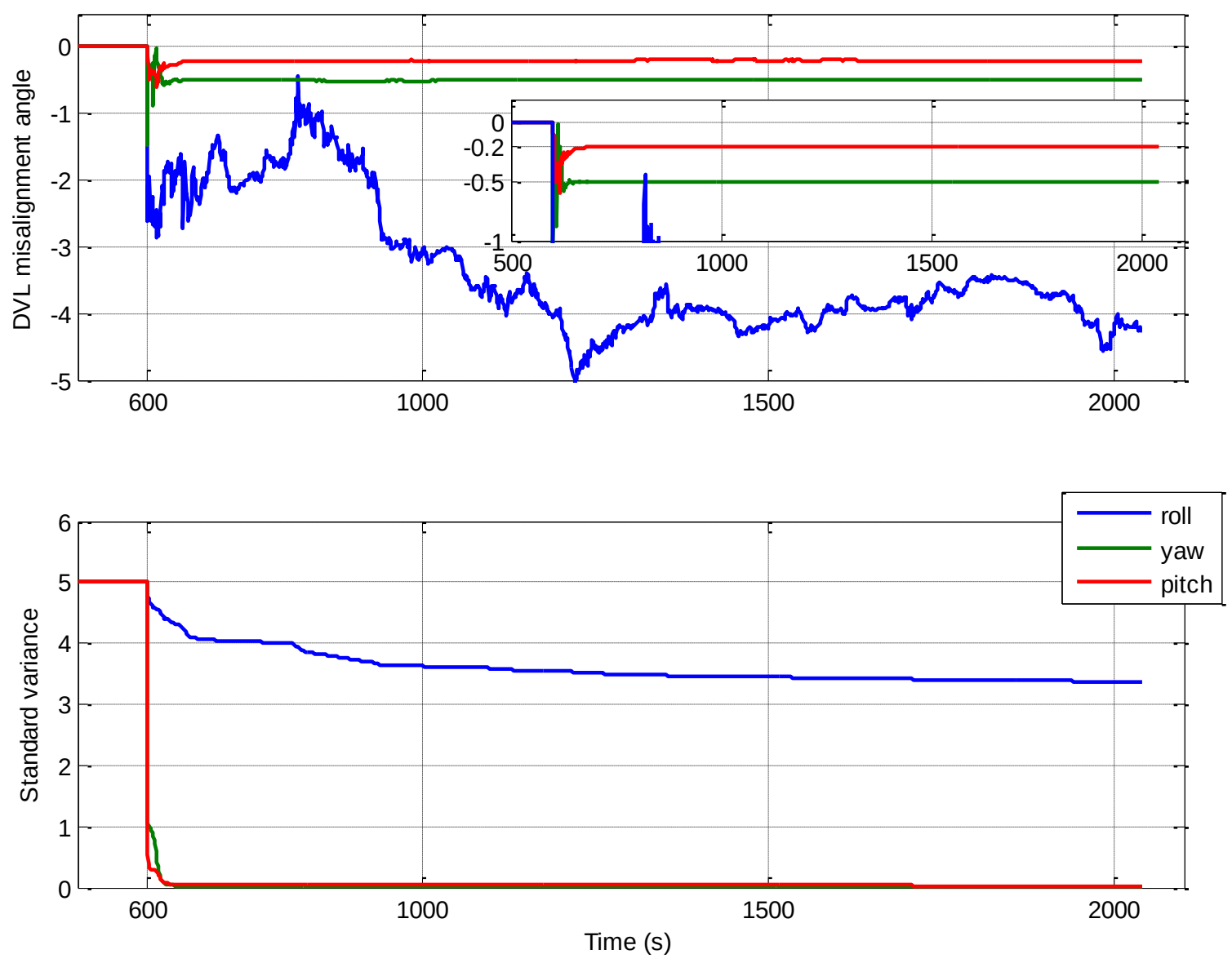

Figure 14. DVL misalignment angle estimate by EKF in 2D run (unit: °).

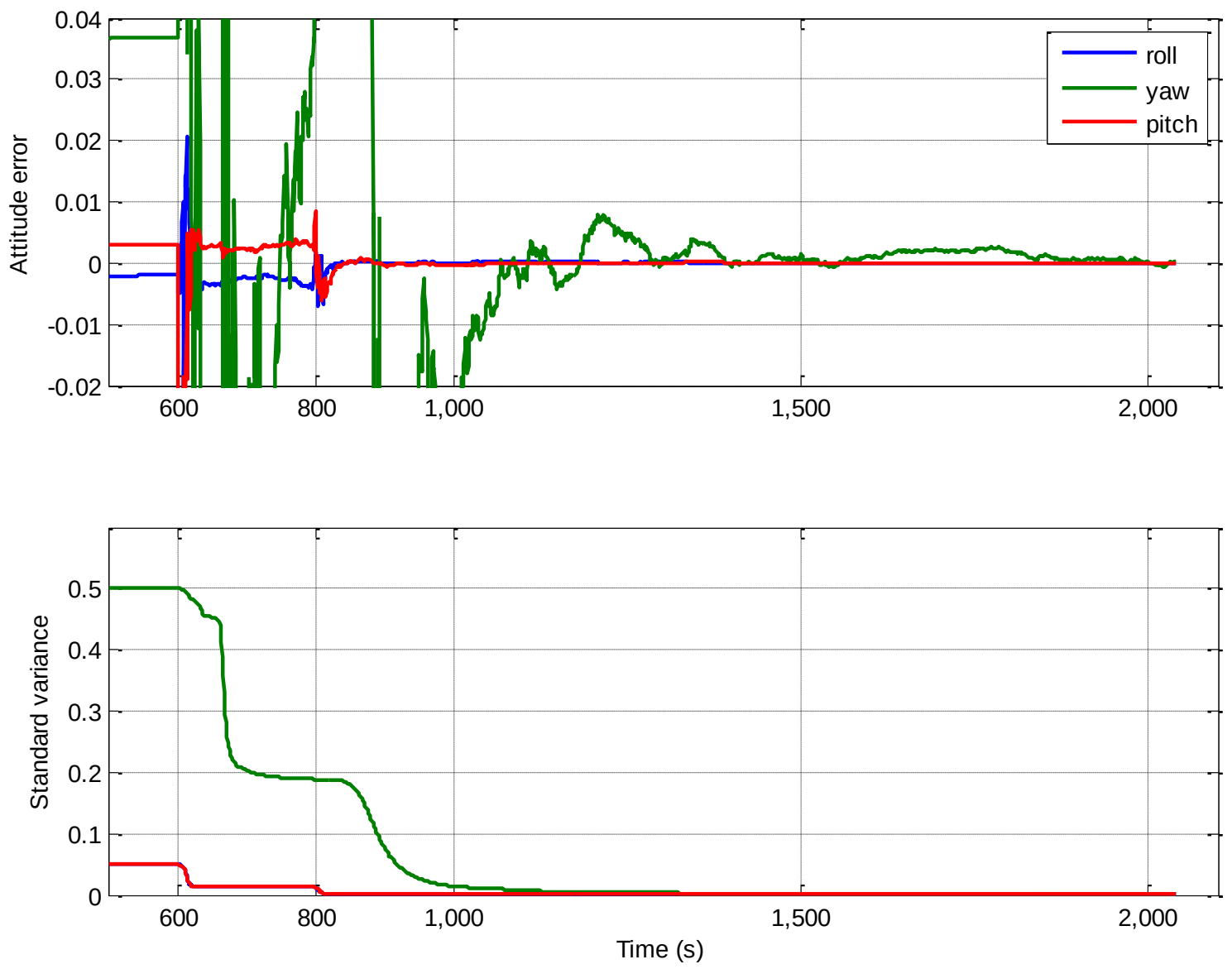

Figure 15. Attitude error and standard variance by EKF in 2D run (unit: °).

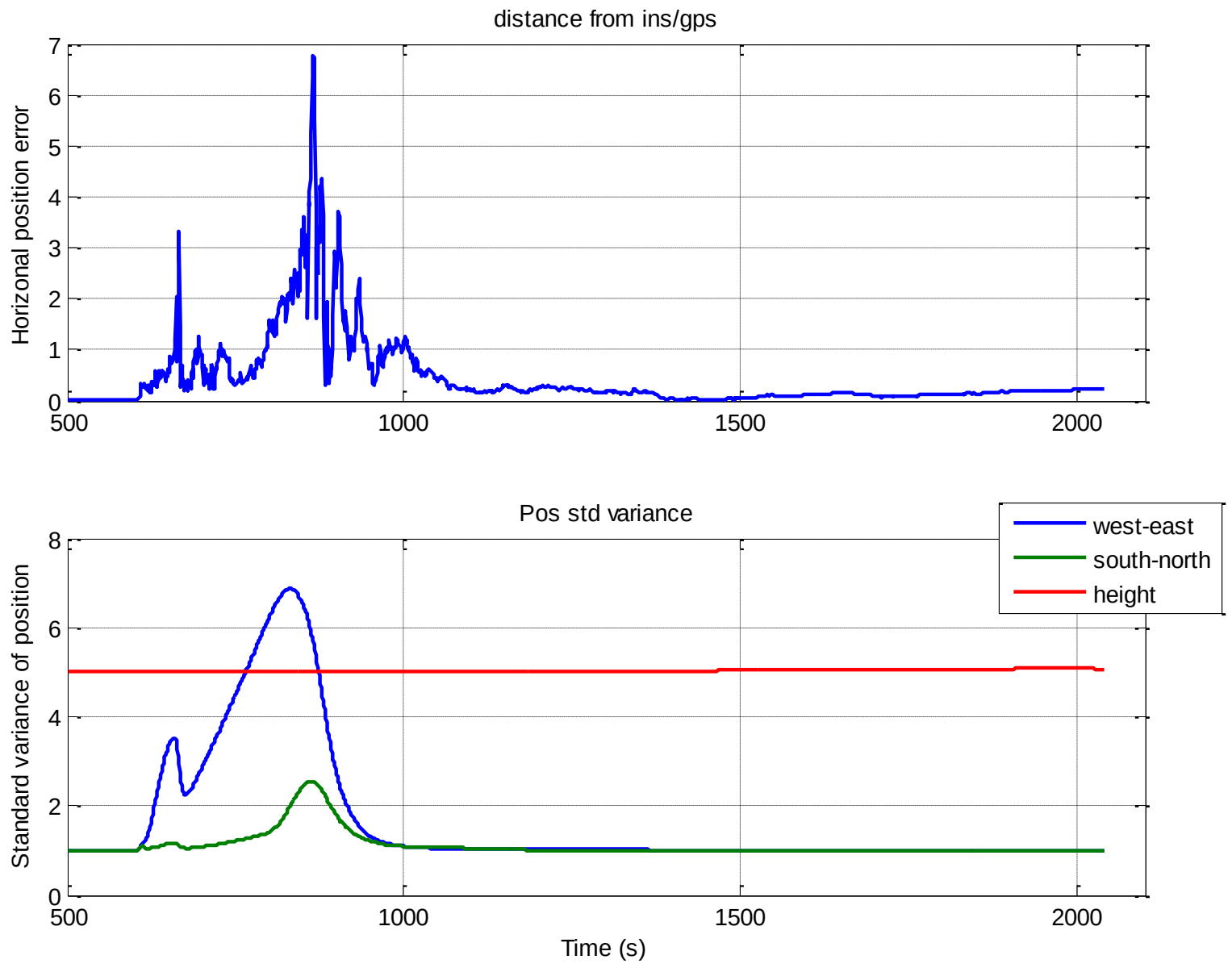

Figure 16. Horizontal position error and standard variance of position estimate by EKF in 2D run (unit: metre).

4. CONCLUSIONS. The development of commercial gyroscope/accelerometer and Doppler sensors has enabled significant improvements to underwater navigation, especially in unexplored areas without artificial beacons. This paper addresses combined IMU/DVL underwater navigation from the viewpoint of the control system. We show by analysis that the combined system is observable under moderate motion conditions. The DVL parameters, such as the scale factor and misalignment angles, especially can be *in-situ* calibrated without relying on additional external GPS or acoustic beacons. The IMU bias errors can also be effectively estimated, aided by the DVL measurement. These benefits are promising to enable a fully autonomous underwater navigation. We carried out numerical simulations and used a practical EKF to integrate IMU and DVL information. The simulation results accord very well with our analytic conclusions. The essential result also applies to IMU/Doppler laser airborne applications. The analytic result is important and a necessary step to ultimately solve the challenging problem of *in-situ* IMU/DVL self-calibration. It also provides useful guidance for field test planning and implementation. High quality field test data will be collected in the future and used to verify the proposed IMU/DVL integration scheme in this paper.

## FINANCIAL SUPPORT

This work was supported in part by the Fok Ying Tung Foundation (grant number 131061) and National Natural Science Foundation of China (grant number 61174002, 61422311).

## REFERENCES

Black, H. D. (1964). A passive system for determining the attitude of a satellite. *AIAA Journal*, **2,** 1350-1351.

Chen, C.-T. (1999). *Linear System Theory and Design*, Rinehart and Winston, Inc.

Groves, P. D. (2008). *Principles of GNSS, Inertial, and Multisensor Integrated Navigation Systems*, Artech House, Boston and London.

Hegren S, Ø. and Berglund, E. (2009). Doppler Water-Track Aided Inertial Navigation for Autonomous

Underwater Vehicle. *OCEANS 2009 - EUROPE.*

Jalving, B., Gade, K., Svartveit, K., Willumsen, A. and Sorhagen, R. (2004). DVL Velocity Aiding in the HUGIN 1000 Integrated Inertial Navigation System. *Modeling, Identification and Control,* **25,** 223-236.

Joyce, T. M. (1989). On in situ calibration of shipboard ADCPs. *Journal of Atmospheric and Oceanic Technology,* **6,** 169–172.

Kinsey, J. C., Eustice, R. M. and Whitcomb, L. L. (2006). A Survey of Underwater Vehicle Navigation: Recent Advances and New Challenges. *IFAC Conference of Manoeuvering and Control of Marine Craft.*

Kinsey, J. C. and Whitcomb, L. L. (2006). Adaptive Identification on the Group of Rigid-Body Rotations and its Application to Underwater Vehicle Navigation. *IEEE Transactions on Robotics,* **23,** 124-136.

Kinsey, J. C. and Whitcomb, L. L. (2007). In Situ Alignment Calibration of Attitude and Doppler Sensors for Precision Underwater Vehicle Navigation: Theory and Experiment. *IEEE Journal of Oceanic Engineering,* **32,** 286-299.

Li, W., Tang, K., Lu, L. and Wu, Y. (2013). Optimization-based INS in-motion alignment approach for underwater vehicles. *Optik - International Journal for Light Electron Optics, in press*.

Savage, P. G. (2007). *Strapdown Analytics,* Strapdown Analysis.

Shuster, M. D. and Oh, S. D. (1981). Three-axis attitude determination from vector observations. *Journal of Guidance, Control, and Dynamics,* 4**,** 70-77.

Silson, P. M. G. (2011). Coarse Alignment of a Ship's Strapdown Inertial Attitude Reference System Using Velocity Loci. *IEEE Trans. on Instrumentation and Measurement,* **60,** 1930-1941.

Tang, K., Wang, J., Li, W. and Wu, W. (2013). A Novel INS and Doppler Sensors Calibration Method for Long Range Underwater Vehicle Navigation. *Sensors,* **13,** 14583-14600.

Tang, Y., Wu, Y., Wu, M., Wu, W., Hu, X. and Shen, L. (2009). INS/GPS integration: global observability analysis. *IEEE Trans. on Vehicular Technology,* **58,** 1129-1142.

Titterton, D. H. and Weston, J. L. (2004). *Strapdown Inertial Navigation Technology,* the Institute of Electrical Engineers, London, United Kingdom, 2nd Ed.

Troni, G., Kinsey, J. C., Yoerger, D. R. and Whitcomb, L. L. (2012). Field Performance Evaluation of New Methods for In-Situ Calibration of Attitude and Doppler Sensors for Underwater Vehicle Navigation. *IEEE International Conference on Robotics and Automation.*

Troni, G. and Whitcomb, L. L. (2010). New Methods for In-Situ Calibration of Attitude and Doppler Sensors for Underwater Vehicle Navigation: Preliminary Results. *OCEANS 2010.* Seattle, WA.

Whitcomb, L., Yoerger, D. and Singh, H. (1999). Advances in Doppler-based navigation of underwater robotic vehicles. *IEEE International Conference on Robotics and Automation.*

Wu, Y. and Pan, X. (2013a). Velocity/Position Integration Formula (I): Application to In-flight Coarse Alignment. *IEEE Transactions on Aerospace and Electronic Systems,* **49,** 1006-1023.

Wu, Y. and Pan, X. (2013b). Velocity/Position Integration Formula (II): Application to Strapdown Inertial Navigation Computation. *IEEE Transactions on Aerospace and Electronic Systems,* **49,** 1024-1034.

Wu, Y., Wang, J. and Hu, D. (2014). A New Technique for INS/GNSS Attitude and Parameter Estimation Using Online Optimization. *IEEE Transactions on Signal Processing,* **62,** 2642 - 2655.

Wu, Y., Wu, M., Hu, X. and Hu, D. (2009). Self-calibration for Land Navigation Using Inertial Sensors and Odometer: Observability Analysis. *AIAA Guidance, Navigation and Control Conference.* Chicago, Illinois, USA.

Wu, Y., Zhang, H., Wu, M., Hu, X. and Hu, D. (2012). Observability of SINS Alignment: A Global Perspective. *IEEE Transactions on Aerospace and Electronic Systems,* **48,** 78-102.

## APPENDIX

*Lemma 1* (Black, 1964; Shuster and Oh, 1981): For any two linearly independent vectors, if their coordinates in two arbitrary frames are given, then the attitude matrix between the two frames can be determined.

*Lemma 2* (Wu et al., 2012): Given $m$ known points $\mathbf{a}_i$, $i = 1, 2, \ldots, m$, in three-dimensional space satisfying $\|\mathbf{a}_i - \mathbf{x}\| = r$, where $\mathbf{x}$ is an unknown point, $r$ is a positive scalar. If points $\mathbf{a}_i$ do not lie in any common plane, $\mathbf{x}$ has a unique solution.